\documentclass[runningheads]{llncs}
\usepackage{graphicx}

\usepackage{tikz}
\usepackage{comment}
\usepackage{amsmath,amssymb} 
\usepackage{color}
\usepackage{xcolor}
\usepackage{hhline}
\usepackage[normalem]{ulem}
\usepackage{multirow}
\usepackage{arydshln}
\usepackage{bbding}
\usepackage{amssymb}
\usepackage{hyperref}
\usepackage{wrapfig}
\usepackage{adjustbox}
\usepackage{pifont} 
\usepackage[accsupp]{axessibility}  

\newcommand{\cmark}{\ding{51}}%
\newcommand{\xmark}{\ding{55}}%

\newlength\savewidth\newcommand\shline{\noalign{\global\savewidth\arrayrulewidth
  \global\arrayrulewidth 1pt}\hline\noalign{\global\arrayrulewidth\savewidth}}
\newcommand{\tablestyle}[2]{\setlength{\tabcolsep}{#1}\renewcommand{\arraystretch}{#2}\centering\footnotesize}
\makeatletter\renewcommand\paragraph{\@startsection{paragraph}{4}{\z@}
  {.5em \@plus1ex \@minus.2ex}{-.5em}{\normalfont\normalsize\bfseries}}\makeatother
\newcommand{\tabincell}[2]{\begin{tabular}{@{}#1@{}}#2\end{tabular}}

\begin{document}
\pagestyle{headings}
\mainmatter
\def\ECCVSubNumber{4824}  

\title{Semantic-Aware Fine-Grained Correspondence} 

\titlerunning{Semantic-Aware Fine-Grained Correspondence}
%

\newcommand*\samethanks[1][\value{footnote}]{\footnotemark[#1]}
\author{Yingdong Hu\inst{1} \and
Renhao Wang\inst{1} \and
Kaifeng Zhang\inst{1} \and
Yang Gao\inst{1,2}\thanks{Corresponding author.}
}

%
\authorrunning{Y. Hu et al.}
%
\institute{Tsinghua University \and Shanghai Qi Zhi Institute\\
\email{\{huyd21,wangrh21,zhangkf19\}@mails.tsinghua.edu.cn}\\
\email{\{gaoyangiiis\}@tsinghua.edu.cn}}

\maketitle

\begin{abstract}
Establishing visual correspondence across images is a challenging and essential task. Recently, an influx of self-supervised methods have been proposed to better learn representations for visual correspondence. However, we find that these methods often fail to leverage semantic information and over-rely on the matching of low-level features. In contrast, human vision is capable of distinguishing between distinct objects as a pretext to tracking. Inspired by this paradigm, we propose to learn semantic-aware fine-grained correspondence. Firstly, we demonstrate that semantic correspondence is implicitly available through a rich set of image-level self-supervised methods. We further design a pixel-level self-supervised learning objective which specifically targets fine-grained correspondence. For downstream tasks, we fuse these two kinds of complementary correspondence representations together, demonstrating that they boost performance synergistically. Our method surpasses previous state-of-the-art self-supervised methods using convolutional networks on a variety of visual correspondence tasks, including video object segmentation, human pose tracking, and human part tracking. Code is available at \url{https://github.com/Alxead/SFC}.
\end{abstract}

\section{Introduction}

Correspondence is considered one of the most fundamental problems in computer vision. At their core, many tasks require learning visual correspondence across space and time, such as video object segmentation~\cite{oh2019video,cheng2021rethinking,wang2019fast}, object tracking~\cite{liu2008sift,henriques2014high,bertinetto2016fully,li2018high}, and optical flow estimation~\cite{dosovitskiy2015flownet,ilg2017flownet,ranjan2017optical,sun2018pwc,teed2020raft}. Despite its importance, prior art in visual correspondence has largely relied on supervised learning~\cite{wang2013learning,held2016learning,valmadre2017end}, which requires costly human annotations that are difficult to obtain at scale. Other works rely on weak supervision from methods like off-the-shelf optical flow estimators, or synthetic training data, which lead to generalization issues when confronted with the long-tailed distribution of real world images.

Recognizing these limitations, many recent works~\cite{vondrick2018tracking,wang2019learning,li2019joint,lai2019self,lai2020mast,jabri2020walk,xu2021rethinking} are exploring self-supervision to learn robust representations of spatiotemporal visual correspondence. Aside from creatively leveraging self-supervisory signals across space and time, these works generally share a critical tenet: evaluation on label propagation as an indication of representation quality. Given label information, such as segmentation labels or object keypoints, within an initial frame, the goal is to propagate these labels to subsequent frames based on correspondence. 

\begin{figure}[t]
\begin{center}
\includegraphics[width=.8\columnwidth]{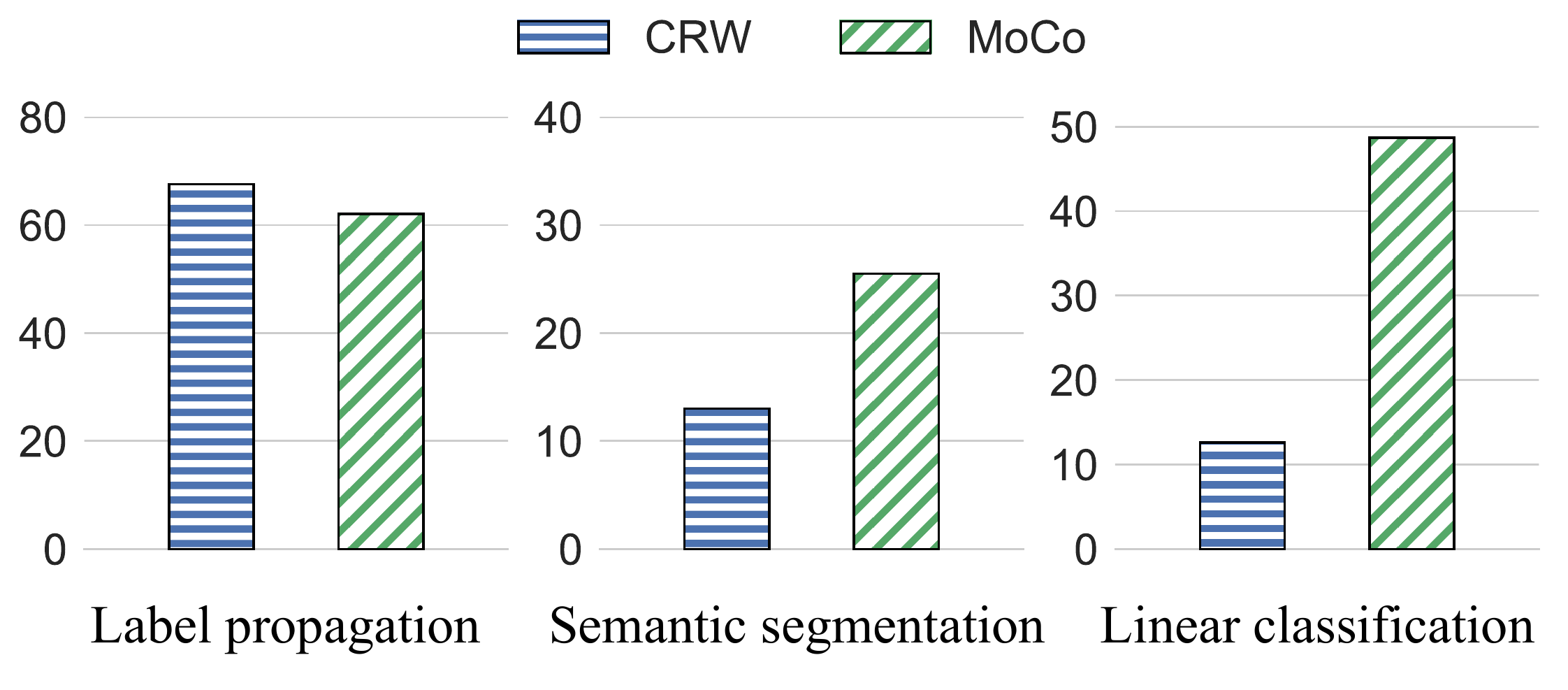}
\caption{We compare Contrastive Random Walk (CRW)~\cite{jabri2020walk} and MoCo~\cite{he2020momentum} on three different downstream tasks. CRW surpasses MoCo on the label propagation task, but is dramatically outperformed by MoCo on semantic segmentation and image classification (details in Appendix~\ref{sec:semantic_and_linear}).}
\label{fig:main_goal}
\end{center}
\vspace{-3em}
\end{figure}

Let us briefly consider the human visual system and how it performs tracking. Many works have argued that our ability to track objects is rooted in our ability to distinguish and understand differences between said objects \cite{grabner2010tracking,gould2007peripheral}. We have rough internal models of different objects, which we adjust by attending more closely to local locations that require fine-grained matching \cite{zhang2020human,ballard2021hierarchical}. In other words, both high-level semantic information and low-level fined-grained information play an important role in correspondence for real visual systems. 

However, when we examine current self-supervised correspondence learning methods, we find them lacking under this paradigm. These methods often overprioritize performance on the label propagation task, and fail to leverage semantic information as well as humans can. In particular, when representations obtained under these methods are transferred to other downstream tasks which require a deeper semantic understanding of images, performance noticeably suffers (see Figure~\ref{fig:main_goal}). We show that label propagation and tracking-style tasks rely on frame-to-frame differentiation of low-level features, a kind of ``shortcut'' exploited by the contrastive-based self-supervised algorithms developed so far. Thus, representations learned via these tasks contain limited semantic information, and underperform drastically when used in alternative tasks.

To this end, we propose Semantic-aware Fine-grained Correspondence (SFC), which simultaneously takes into account semantic correspondence and fine-grained correspondence. 
Firstly, we find that current image-level self-supervised representation learning methods e.g. MoCo~\cite{he2020momentum} force the mid-level convolutional features to implicitly capture correspondences between similar objects or parts. Second, we design an objective which learns high-fidelity representations of fine-grained correspondence (FC). We do this by extending prior image-level loss functions in self-supervised representation learning to a dense paradigm, thereby encouraging local feature consistency. Crucially, FC does not use temporal information to learn this low-level correspondence, but our ablations show that this extension alone makes our model competitive with previous methods relying on temporal signals in large video datasets for pretraining. 

\begin{figure*}[t]
\begin{center}
\includegraphics[width=\linewidth]{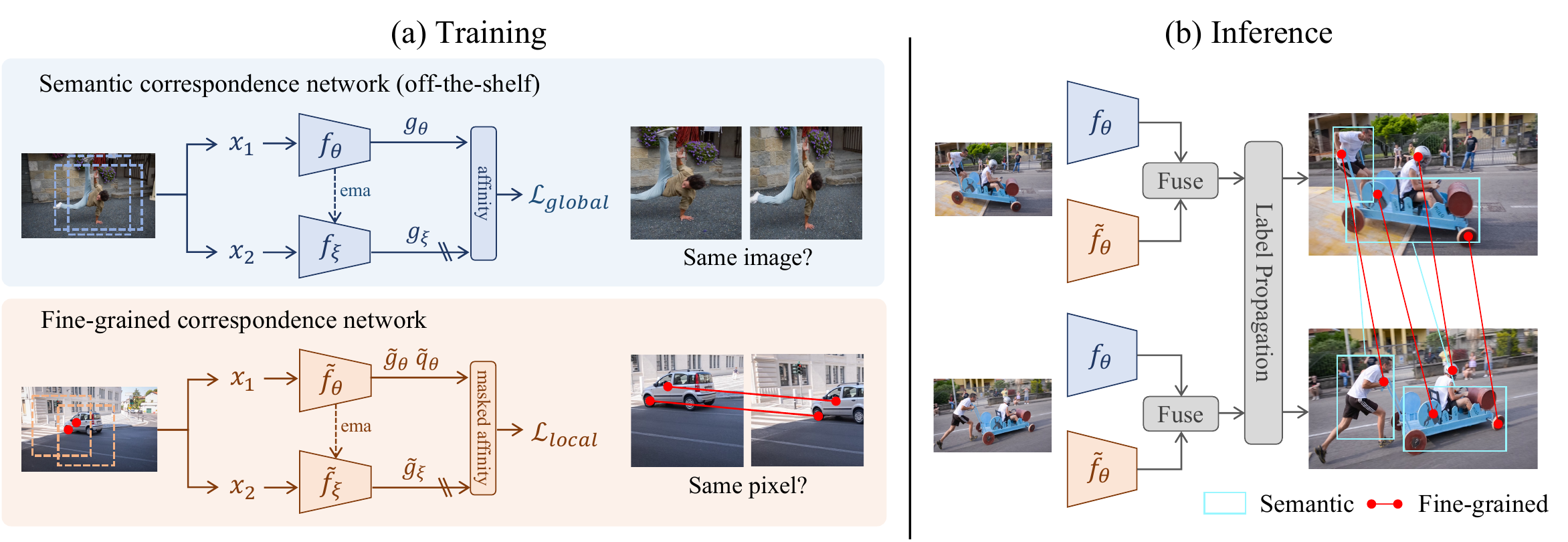}
\caption{\textbf{Overview of Semantic-aware Fine-grained Correspondence learning framework.} By maximizing agreement between positive (similar) image pairs, convolutional representations capture semantic correspondences between similar objects implicitly. By encouraging the spatially close local feature vectors to be consistent, model can learn fine-grained correspondence explicitly. For downstream task, we utilize two kinds of correspondence together to achieve complementary effects.}
\label{fig:model}
\end{center}
\vskip -0.3 in
\end{figure*}

Prior works~\cite{wang2021dense,xie2021propagate} have shown that image-level self-supervision can further facilitate the dense self-supervision in a multitask framework. However, we surprisingly find that our fine-grained training objective and image-level semantic training objectives are inconsistent: each of them requires the model to encode conflicting information about the image, leading to degradation in performance when used in conjunction. We hypothesize that it is necessary to have two independent models, and propose a late fusion operation to combine separately pretrained semantic correspondence and fine-grained correspondence feature vectors. Figure~\ref{fig:model} overviews the proposed method. Through our ablations, we categorically verify that low-level fine-grained correspondence and high-level semantic correspondence are complementary, and indeed orthogonal, in the benefits they bring to self-supervised representation learning. The main contributions of our work are as follows:

\begin{itemize}
    \item We propose to learn semantic-aware fine-grained correspondence (SFC), while most previous works consider and improve the two kinds of correspondence separately.
    \item We design a simple and effective self-supervised learning method tailored for low-level fine-grained correspondence. Despite using static images and discarding temporal information, we outperform previous methods trained on large-scale video datasets.
    \item Late fusion is an effective mechanism to prevent conflicting image-level and fine-grained training objectives from interfering with each other.
    \item Our full model (SFC) sets the new state-of-the-art for self-supervised approaches using convolutional networks on various video label propagation tasks, including video object segmentation, human pose tracking, and human part tracking.
\end{itemize}

\section{Related Work}

\noindent \textbf{Self-Supervised Representation Learning} \quad Self-supervised representation learning has gained popularity because of its ability to avoid the cost of annotating large-scale datasets. Specifically, methods using instance-level pretext tasks have recently become dominant components in self-supervised learning for computer vision ~\cite{wu2018unsupervised,oord2018representation,henaff2020data,bachman2019learning,tian2020contrastive,he2020momentum,chen2020improved,misra2020self,chen2020simple,caron2020unsupervised,gordon2020watching,purushwalkam2020demystifying}. Instance-level discrimination aims to pull embeddings of augmented views of the same image (positive pairs) close to each other, while trying to push away embeddings from different images (negative pairs). Recently, some works~\cite{grill2020bootstrap,chen2021exploring} have discovered that even without negative pairs, self-supervised learning can exhibit strong performance. 

To obtain better transfer performance to dense prediction tasks such as object detection and semantic segmentation, other works~\cite{pinheiro2020unsupervised,xie2021propagate,wang2021dense} explore pretext tasks at the pixel level for representation learning. But empirically, we find that these methods fail to leverage fine-grained information well. Our fine-grained correspondence network (FC) is most closely related to PixPro~\cite{xie2021propagate} which obtains positive pairs by extracting features from the same pixel through two asymmetric pipelines. Both FC and PixPro can be seen as dense versions of BYOL~\cite{grill2020bootstrap}, but the two methods have completely different goals. FC has many design choices tailored for correspondence learning: FC preserves spatial sensitivity by avoiding entirely a pixel propagation module which introduces a certain smoothing effect. Furthermore, we discard color augmentation and use higher resolution feature maps, as we find both modifications are beneficial to the fine-grained correspondence task. Finally, FC can achieve competitive performance to predominant approaches, with compelling computational and data efficiency. In contrast, the transfer performance of PixPro on the correspondence task is far behind its instance-level counterpart~\cite{grill2020bootstrap} and our FC.

We note that DINO~\cite{caron2021emerging}, a self-supervised Vision Transformer (ViT)~\cite{dosovitskiy2020image}, exhibits surprisingly strong correspondence properties and competitive performance on DAVIS-2017 benchmark. We speculate that the success of DINO on this task is attributed to the architecture of ViT and much more computation.

\smallskip\noindent \textbf{Self-Supervised Correspondence Learning} \quad Recently, numerous approaches have also been developed for correspondence learning in a self-supervised manner~\cite{vondrick2018tracking,wang2019learning,wang2019unsupervised,lai2019self,lai2020mast,li2019joint,jabri2020walk,xu2021rethinking}. The key idea behind a number of these methods~\cite{vondrick2018tracking,lai2019self,lai2020mast} is to propagate the color of one frame in a video to future frames. TimeCycle~\cite{wang2019learning} relies on a cycle-consistent tracking pretext task. Along this line, CRW~\cite{jabri2020walk} cast correspondence as pathfinding on a space-time graph, also using cycle-consistency as a self-supervisory signal. VFS~\cite{xu2021rethinking} propose to learn correspondence implicitly by performing image-level similarity learning. Despite the success of these methods, they all rely heavily on temporal information from videos as the core form of self-supervision signal. In our work, we demonstrate that representations with good space-time correspondence can be learned even without videos. Moreover, our framework is an entirely alternative perspective on correspondence learning, which can be flexibly adapted with other video-based methods to further improve performance.

\smallskip\noindent \textbf{Semantic Correspondence} \quad We borrow the notion of semantic correspondence from literature~\cite{rocco2017convolutional,rocco2018end,rocco2018neighbourhood,min2019hyperpixel,min2020learning,liu2020semantic,min2021convolutional,truong2020glu}, which aim to establish dense correspondences across images depicting different instances of the same object categories. Evaluation of these methods exists solely on image datasets with keypoint annotations, which can be more forgiving and translates poorly to the real world. Our semantic correspondence is evaluated on video, which we argue is a much more realistic setting for correspondence. In addition, many supervised semantic correspondence approaches~\cite{chen2018deep,min2019hyperpixel,huang2019dynamic,liu2020semantic} adopt a CNN pre-trained on image classification as their frozen backbone, but we explore a self-supervised pre-trained backbone as an alternative.

\section{Method}

While our framework is compatible with a wide array of contemporary self-supervised representation methods, we demonstrate its efficacy with two recent approaches: MoCo~\cite{he2020momentum} and BYOL~\cite{grill2020bootstrap}, which are reviewed in Section~\ref{sec:background}. Next, in Section~\ref{sec:semantic}, we argue that image-level methods implicitly learn high-level semantic correspondence. In Section~\ref{sec:fc}, we propose our framework to improve fine-grained correspondence learning. Finally, in Section~\ref{sec:combined}, we show how to unify these two complementary forms of correspondence to improve performance on video label propagation tasks.

\subsection{Background}
\label{sec:background}

In image-level self-supervised representation learning, we seek to minimize a distance metric between two random augmentations $\mathbf{x}_{1}$ and $\mathbf{x}_{2}$ of a single image $\mathbf{x}$. One popular framework for doing this is contrastive learning~\cite{hadsell2006dimensionality}. 

Formally, two augmented views $\mathbf{x}_{1}$ and $\mathbf{x}_{2}$ are fed into an online encoder and target encoder respectively, where each encoder consists of a backbone $\mathbf{f}$ (e.g. ResNet), and a projection MLP head $\mathbf{g}$. The $l_2$-normalized output global feature vectors for $\mathbf{x}_{1}$ and $\mathbf{x}_{2}$ can be represented as $\mathbf{z}_{1} \triangleq \mathbf{g}_{\boldsymbol{\theta}}(\mathbf{f}_{\boldsymbol{\theta}}(\mathbf{x}_{1}))$ and $\mathbf{z}_{2} \triangleq \mathbf{g}_{\boldsymbol{\xi}}(\mathbf{f}_{\boldsymbol{\xi}}(\mathbf{x}_{2}))$, where $\boldsymbol{\theta}$ and $\boldsymbol{\xi}$ are parameters of the two respective networks. Let the negative features obtained from $K$ different images be represented by the set $\mathcal{S} = \{\mathbf{s}_{1}, \mathbf{s}_{2}, \dots, \mathbf{s}_{K} \}$. Then contrastive learning uses the InfoNCE~\cite{oord2018representation} to pull $\mathbf{z}_{1}$ close to $\mathbf{z}_{2}$ while pushing it away from negative features:

\begin{equation}
\mathcal{L}_{\text{InfoNCE}} = -\log \frac{\exp (\mathbf{z}_{1} {\cdot} \mathbf{z}_{2} / \tau)}{\exp (\mathbf{z}_{1} {\cdot} \mathbf{z}_{2} / \tau) + \sum_{k = 1}^K \exp(\mathbf{z}_{1} {\cdot} \mathbf{s}_{k} / \tau)}
\end{equation} where $\tau$ is the temperature hyperparameter. While numerous methods have been explored to construct the set of negative samples $\mathcal{S}$, we choose MoCo~\cite{he2020momentum} for obtaining semantic correspondence representations, which achieves this goal via a momentum-updated queue. In particular, the target encoder's parameters $\boldsymbol{\xi}$ are the exponential moving average of the online parameters $\boldsymbol{\theta}$:

\begin{equation}
    \boldsymbol{\xi} \leftarrow  m \boldsymbol{\xi} + (1-m)\boldsymbol{\theta}, \qquad m \in [0, 1]
\end{equation} where $m$ is the exponential moving average parameter.

Some recent works~\cite{grill2020bootstrap,chen2021exploring} show that it is not necessary to use negative pairs to perform self-supervised representation learning. One such method, BYOL~\cite{grill2020bootstrap}, relies on an additional prediction MLP head $\mathbf{q}_{\boldsymbol{\theta}}$ to transform the output of online encoder $\mathbf{p}_{1} \triangleq \mathbf{q}_{\boldsymbol{\theta}}(\mathbf{z}_{1}) $. The contrastive objective then reduces to simply minimizing the negative cosine distance between the predicted features $\mathbf{p}_{1}$ and the features obtained from the target encoder $\mathbf{z}_{2}$ ($l_2$-normalized):

\begin{equation}
    \mathcal{L}_{\text{global}} = - \frac{\left<\mathbf{p}_1, \mathbf{z}_2 \right >}{\| \mathbf{p}_1\|_{2}\cdot \| \mathbf{z}_2\|_{2}}
\end{equation}

Note again that MoCo and BYOL bear striking similarities in their formulation and training objectives. 
In the following section, we hypothesize that such similarities in frameworks lead to similarities in types of features learned. In particular, we claim that image-level representations in general contain information about semantic correspondences.

\subsection{Semantic Correspondence Learning}
\label{sec:semantic}

Representations learned by current self-supervised correspondence learning methods may contain limited semantic information. To make the representations more neurophysiologically intuitive, we add the crucially missing semantic correspondence learning into our method. Recent image-level self-supervised methods learn representations by imposing invariances to various data augmentations. Two random crops sampled from the same image, followed by strong color augmentation~\cite{chen2020simple} are considered as positive pairs. The augmentation significantly changes the visual appearance of the image 
but keeps the semantic meaning unchanged. The model can match positive pairs by attending only to the essential part of the representation, while ignoring other non-essential variations. As a result, different images with similar visual concepts are grouped together, inducing a latent space with rich semantic information~\cite{van2020scan,chen2020intriguing,van2021revisiting}. This is evidenced by the results shown in Figure~\ref{fig:main_goal}, where MoCo~\cite{he2020momentum} achieve high performance on tasks that require a deeper semantic understanding of images.
Moreover, previous works~\cite{long2014convnets,xu2021rethinking} demonstrate that correspondence naturally emerges in the middle-level convolutional features. Thus we conclude that current self-supervised representation methods can implicitly learn semantic correspondence well. 

We utilize one approach, MoCo~\cite{he2020momentum}, in our downstream correspondence task. In particular, only the pre-trained online backbone $\mathbf{f}_{\boldsymbol{\theta}}$ is retained, while all other parts of the network, including the online projection head $\mathbf{g}_{\boldsymbol{\theta}}$ and target encoder $\mathbf{f}_{\boldsymbol{\xi}}, \mathbf{g}_{\boldsymbol{\xi}}$, are discarded. We use $\mathbf{f}_{\boldsymbol{\theta}}$ to encode each image as a semantic correspondence feature map: $\mathbf{F} = \mathbf{f}_{\boldsymbol{\theta}}(\mathbf{x}) \in \mathbb{R}^{H \times W \times C_{s}} $, where $H$ and $W$ are spatial dimensions.
Note also that we can adjust the size of the feature map by changing the stride of residual blocks, offering additional flexibility in the scale of semantic information we wish to imbue our representations with.

Finally, we comment that the emergent mid-level feature behavior extends readily to MoCo, and moreover also to other self-supervised methods like BYOL~\cite{grill2020bootstrap} and SimCLR~\cite{chen2020simple}, as the encoders for all such methods are based on ResNet-style architectures. We can thus flexibly swap out the semantic correspondence backbone for any of these image-level self-supervised representations.

\subsection{Fine-grained Correspondence Learning}
\label{sec:fc}
Only considering semantic information is not enough for correspondence learning, which often requires analyzing low-level variables such as object edge, pose, articulation, precise location and so on. Like most previous self-supervised methods, we also incorporate low-level fine-grained correspondence in our approach. BYOL-style methods~\cite{grill2020bootstrap} learn their representations by directly maximizing the similarity of two views of one image (positive pairs) in the feature space. This paradigm naturally connects with our intuitive understanding of correspondence: similar objects, parts and pixels should have similar representations. We are thus inspired to generalize this framework to a dense paradigm to learn fine-grained correspondence specifically. 

At a high level, we learn our embedding space by pulling local feature vectors belonging to the same spatial region close together. Specifically, given two augmented views $\mathbf{x}_1$ and $\mathbf{x}_{2}$ of one image,  we extract their dense feature maps $\mathbf{F}_{1} \triangleq \tilde{\mathbf{f}}_{\boldsymbol{\theta}}(\mathbf{x}_{1}) \in \mathbb{R}^{H \times W \times C_{f}} $ and $\mathbf{F}_{2} \triangleq \tilde{\mathbf{f}}_{\boldsymbol{\xi}}(\mathbf{x}_{2}) \in \mathbb{R}^{H \times W \times C_{f}}$ by removing the global pooling layer in the encoders. We adopt a ResNet-style backbone, and we can thus reduce the stride of some residual blocks in order to obtain a higher resolution feature map. In addition, to maintain dense 2D feature vectors, we replace the MLPs in the projection head and prediction head with $1 \times 1$ convolution layers. Then we can get dense prediction feature vectors $\mathbf{P}_{1} \triangleq \tilde{\mathbf{q}}_{\theta}(\tilde{\mathbf{g}}_{\theta}(\mathbf{F}_{1})) \in \mathbb{R}^{H \times W \times D}$ and dense projection feature vectors $\mathbf{Z}_{2} \triangleq \tilde{\mathbf{g}}_{\xi}(\mathbf{F}_{2}) \in \mathbb{R}^{H \times W \times D}$. $\mathbf{P}_{1}^{i}$ denotes the local feature vector at the $i$-th position of $\mathbf{P}_{1}$. 
Now, a significant question remains: for a given local feature vector $\mathbf{P}_{1}^{i}$, how can we find its positive correspondence local feature vector in $\mathbf{Z}_{2}$?

\smallskip \noindent \textbf{Positive Correspondence Pairs} \quad Note that after we apply different spatial augmentations (random crop) to the two views, the local feature vectors on the two feature maps are no longer aligned. An object corresponding to a local feature vector in one view may even be cropped in another view. Thus, we only consider feature vectors corresponding to the same cropped region (overlapped areas of two views) and define a small spatial neighborhood around each local feature vector. All the local feature vectors in the spatial neighborhood are designated positive samples.

\begin{wrapfigure}{r}{0.45\textwidth}
\vspace{-3em}
\adjustbox{width=\linewidth}{
\includegraphics[width=0.45\textwidth]{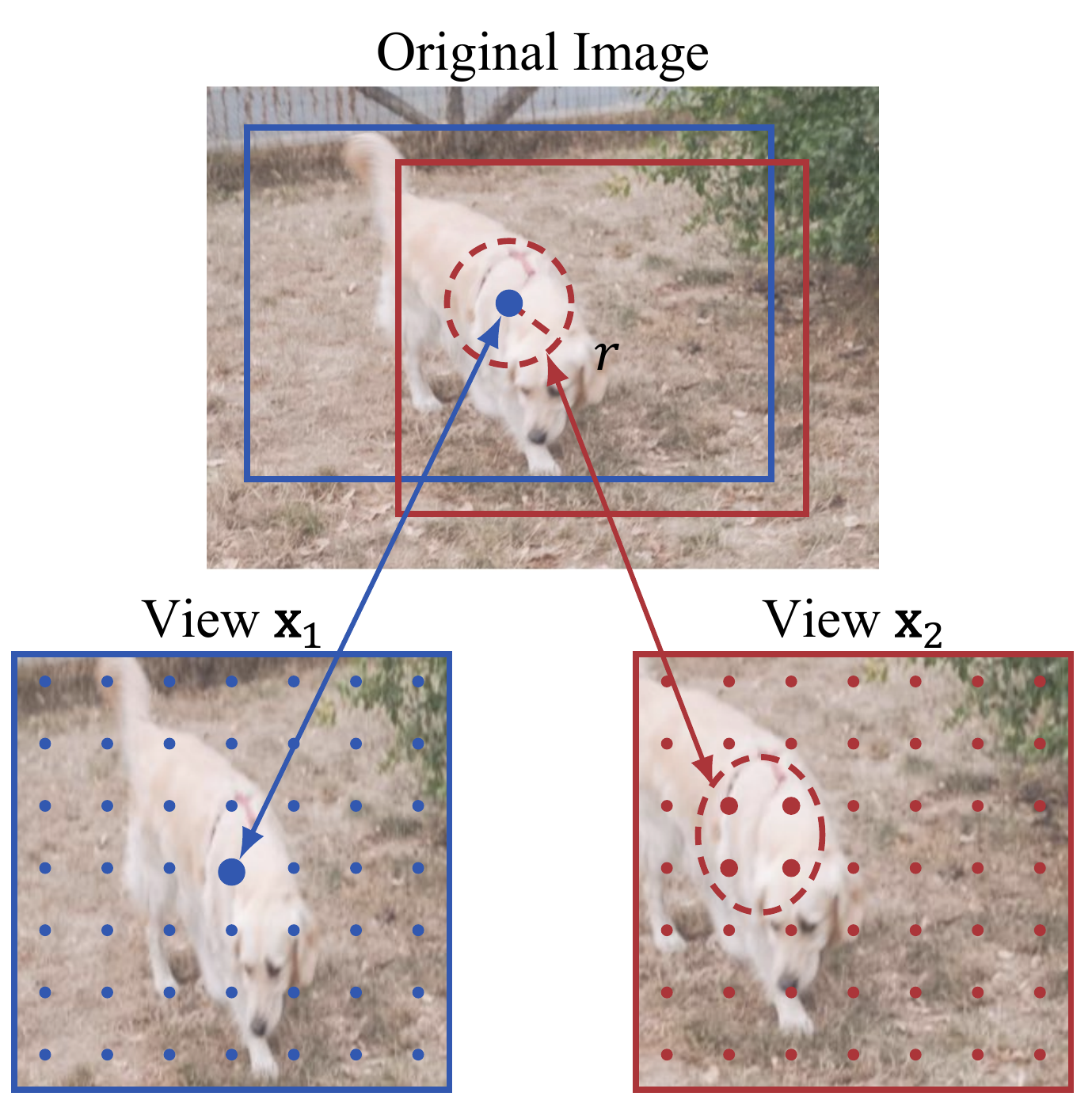}}
\caption{For a feature vector in one view $\mathbf{x}_{1}$, we designate all the feature vectors in view $\mathbf{x}_{2}$ which belonging to the same spatial region as positive pairs.}
\label{fig:fine-grained}
\vspace{-4em}
\end{wrapfigure}

Specifically, we construct a binary positive mask $\mathbf{M} \in \mathbb{R}^{H\cdot W \times H \cdot W} $ by computing the spatial distance between all pairs of local feature vectors with:
\begin{equation}
    \mathbf{M}_{ij}= \begin{cases}
    1 & \operatorname{dist}(\Phi(\mathbf{P}_{1}^{i}), \Phi(\mathbf{Z}_{2}^{j})) \leqslant  r \\
    0 & \operatorname{dist}(\Phi(\mathbf{P}_{1}^{i}), \Phi(\mathbf{Z}_{2}^{j})) >  r \\
    \end{cases}
\end{equation}       $\Phi$ denotes an operation that translates the coordinates of the local feature vector to the original image space. $\operatorname{dist}$ denotes the distance between coordinates of local feature vectors $\mathbf{P}_{1}^{i}$ and $\mathbf{Z}_{2}^{j}$ in the original image space.
$r$ is positive radius, which controls a notion of locality. As we show in the experiment, this is a very important hyperparameter. In summary, all $1$s in the $i$-th row of $\mathbf{M}$ represent the local feature vectors in $\mathbf{Z}_{2}$ which are positive samples of the $i$-th vector in $\mathbf{P}_{1}$. This process is illustrated in Figure~\ref{fig:fine-grained}.

\smallskip \noindent \textbf{Learning Objectives} \quad We construct a pairwise similarity matrix $\mathbf{S}$, where $\mathbf{S} \in \mathbb{R}^{H\cdot W \times H\cdot W} $ with:
\begin{equation}
    \mathbf{S}_{ij} = \operatorname{sim}(\mathbf{P}_{1}^{i}, \mathbf{Z}_{2}^{j})
\end{equation}
$\operatorname{sim}(\mathbf{u}, \mathbf{v})=\frac{\left <\mathbf{u}, \mathbf{v}\right>}{\|\mathbf{u}\|_2\cdot \|\mathbf{v}\|_2}$ denotes the cosine similarity between two vectors. We multiply the similarity matrix $\mathbf{S}$ and the positive mask $\mathbf{M}$ to get the masked similarity matrix $\widetilde{\mathbf{S}}=\mathbf{S} \odot \mathbf{M}$.
Finally, the loss function seeks to maximize each element in the masked similarity matrix $\widetilde{\mathbf{S}}$:

\begin{equation}
    \mathcal{L}_{\text{local}} = - \frac{\sum_{i=1}^{H\cdot W}\sum_{j=1}^{H\cdot W} \widetilde{\mathbf{S}}_{ij}}{\sum_{i=1}^{H\cdot W}\sum_{j=1}^{H\cdot W} \mathbf{M}_{ij}}
\end{equation}

\subsection{Fusion of Correspondence Signals}
\label{sec:combined}

To combine semantic correspondence (Sec.~\ref{sec:semantic}) and fine-grained correspondence (Sec.~\ref{sec:fc}) representations, one intuitive approach is simultaneously train with both semantic-level and fine-grained level losses, like~\cite{wang2021dense,xie2021propagate,bai2022point}. However, our investigations reveal that jointly using both these objectives may not be sensible, as the representations fundamentally conflict, in two main ways. 1)\emph{receptive fields}. We find that fine-grained correspondence relies heavily on a higher resolution feature map (see Appendix~\ref{sec:feature_resolution}). But trivially increasing the feature resolution of a semantic-level method like MoCo~\cite{he2020momentum} during training causes performance on the label propagation task to drop a lot. This is because low-level fine-grained information needs small receptive fields while relatively large receptive fields are necessary to encode global high-level semantic information. 2)\emph{data augmentation}. Similar to VFS~\cite{xu2021rethinking}, we find that color augmentation (e.g. color distortion and grayscale conversion) is harmful to learning fine-grained correspondence, since fine-grained correspondence heavily relies on low-level color and texture details. In contrast, image-level self-supervised learning methods learn semantic representations by imposing invariances on various data transformations. 

For example, as seen in the augmentations ablation for SimCLR (Fig. 5 in~\cite{chen2020simple}), removing color augmentation leads to severe performance issues.

We conclude that an end-to-end framework utilizing multiple levels of supervision does not always work, especially when these modes of supervision have different requirements on both the model and data sides (see Sec.~\ref{sec:fusion_strategy} for experimental evidence). We argue it is necessary to decouple the two models, which is consistent with how humans also attend very differently when re-identifying an object’s main body versus its accurate pixel  boundary. 
Inspired by Two-Stream ConvNets\cite{simonyan2014two}, which use a late fusion to combine two kinds of complementary information, and hypercolumns~\cite{hariharan2015hypercolumns}, which effectively leverage information across different layers of CNNs, we implement a similar mechanism to fuse our orthogonal correspondences.

For a given image, suppose we have two networks, one which produces a semantic correspondence feature map $\mathbf{F}_{s} \triangleq \mathbf{f}_{\boldsymbol{\theta}}(\mathbf{x}) \in \mathbb{R}^{H \times W \times C_{s}} $ and one which produces a fine-grained correspondence feature map $\mathbf{F}_{f} \triangleq \tilde{\mathbf{f}}_{\boldsymbol{\theta}}(\mathbf{x}) \in \mathbb{R}^{H \times W \times C_{f}} $. Note that these two feature maps can have different channel dimensions. We consider channel-wise concatenation as a simple and intuitive way to fuse these feature maps:

\begin{equation}
    \mathbf{F} = [\operatorname{L2Norm}(\mathbf{F}_{s}), \lambda \cdot \operatorname{L2Norm}(\mathbf{F}_{f})]
\end{equation}

where $\operatorname{L2Norm}$ denotes an $l_{2}$ normalization of local feature vectors in every spatial location. This ameliorates issues of scale, considering that the two feature maps are obtained under different training objectives which attend to features of different scales. $\lambda$ is a hyperparameter to balance two feature maps. Note that $\mathbf{F}$ also needs to be re-normalized when it is employed in downstream tasks, like label propagation.

\subsection{Implementation Details}
Any off-the-shelf image-level self-supervised pre-trained network can serve as our semantic correspondence backbone. In our implementation, we use MoCo as the default network, with ResNet-50~\cite{he2016deep} as the base architecture and pre-trained on the 1000-class ImageNet~\cite{deng2009imagenet} training set with strong data augmentation. 

As for our fine-grained correspondence network, we use YouTube-VOS~\cite{xu2018youtube} as our pre-training dataset for direct comparison with previous works~\cite{lai2020mast}. It contains $3471$ videos totalling 5.58 hours of playtime, much smaller than Kinetics400~\cite{carreira2017quo} (800 hours). Although Youtube-VOS is a video dataset, we treat it as a conventional image dataset and randomly sample individual frames during training (equivalent to 95k images). Crucially, this discards temporal information and correspondence signals our model would otherwise be able to exploit. We use \emph{cropping-only} augmentation. Following ~\cite{jabri2020walk,li2019joint,wang2019learning}, we adopt ResNet-18 as the backbone. Please see Appendix~\ref{sec:fc_details} for augmentation, architecture and optimization details.

\section{Experiments}

We evaluate the learned representation without fine-tuning on several challenging video propagation tasks involving objects, human pose and parts. We will first introduce our detailed evaluation settings and baselines, then we conduct the comparison with the state-of-the-art self-supervised algorithms. Finally, we perform extensive ablations on different elements for SFC.

\subsection{Experimental Settings}

\subsubsection{Label propagation} Ideally, a model with good space-time correspondence should be able to track an arbitrary user-annotated target object throughout a video. Previous works formulate this kind of tracking task as video label propagation~\cite{wang2019learning,li2019joint,jabri2020walk,xu2021rethinking}. We follow the same evaluation protocol as prior art~\cite{jabri2020walk} for consistency and equitable comparison. At a high level, we use the representation from our pre-trained model as a similarity function. Given the ground-truth labels in the first frame, a recurrent inference strategy is applied to propagate the labels to the rest of the frames. See Appendix~\ref{sec:label_prop_detail} for detailed description.

We compare with state-of-the-art algorithms on DAVIS-2017~\cite{pont20172017}, a widely-used publicly-available benchmark for video object segmentation. To see whether our method can generalize to more visual correspondence tasks, we further evaluate our method on JHMDB benchmark~\cite{jhuang2013towards}, which involves tracking 15 human pose keypoints, and on the Video Instance Parsing (VIP) benchmark~\cite{zhou2018adaptive}, which involves propagating 20 parts of the human body. We use the same settings as ~\cite{jabri2020walk,li2019joint} and report the standard metrics, namely region-based similarity $\mathcal{J}$ and contour-based accuracy $\mathcal{F}$~\cite{perazzi2016benchmark} for DAVIS, probability of a correct pose (PCK) metric~\cite{yang2011articulated} for JHMDB and mean intersection-over-union (IoU) for VIP.

\begin{table*}[t]
  \centering
  \caption{\textbf{Video object segmentation results on the DAVIS-2017 val set.} \emph{Dataset} indicates dataset(s) used for pre-training, including: I=ImageNet, V=ImageNet-VID, C=COCO, D=DAVIS-2017, P=PASCAL-VOC, J=JHMDB. $\star$ indicates that the method uses its own label propagation algorithm.}
  \label{tab:davis}
  \resizebox{\textwidth}{!}{
  \begin{tabular}{lcc|ccc|cc|c}
&  &  & \multicolumn{3}{c|}{DAVIS}  & \multicolumn{2}{c|}{JHMDB} & \multicolumn{1}{c}{VIP}   \\
Method & Supervised & Dataset (Size) & $\mathcal{J\&F}_\textrm{m}$ & $\mathcal{J}_\textrm{m}$ & $\mathcal{F}_\textrm{m}$  & PCK@0.1 & PCK@0.2  & mIoU \\
  \shline
Rand.Init&\xmark& -       & 32.5 & 32.4 & 32.6 & 50.8 & 72.3 & 18.6 \\
Supervised\cite{he2016deep}&\cmark&I (1.28M)  & 66.9 & 64.5 & 69.4 & 59.7 & 81.2 & 38.6 \\
InstDis\cite{wu2018unsupervised}   &\xmark&I (1.28M)  & 66.4 & 63.9 & 68.9 & 58.5 & 80.2 & 32.5 \\
MoCo\cite{he2020momentum} &\xmark&I (1.28M)  & 65.9 & 63.4  & 68.4 & 59.4 & 80.9 & 33.1 \\
SimCLR\cite{chen2020simple} &\xmark&I (1.28M)  & 66.9 & 64.4  & 69.4 & 59.0 & 80.8 & 35.3  \\
BYOL\cite{grill2020bootstrap} &\xmark&I (1.28M)  & 66.5 & 64.0 & 69.0 & 58.8 & 80.9 & 34.8  \\
SimSiam\cite{chen2021exploring}   &\xmark&I (1.28M)  & 67.2 & 64.8 & 68.8 & 59.9 & 81.6 & 33.8 \\
VINCE\cite{gordon2020watching} &\xmark& Kinetics (800 hours) & 65.2 & 62.5 & 67.8 & 58.8 & 80.4 & 35.3 \\
VFS$^{\star}$\cite{xu2021rethinking}   &\xmark& Kinetics (800 hours) & 68.9 & 66.5 & 71.3 & 60.9 & 80.7 & 43.2   \\
\hdashline
DetCo\cite{xie2021detco}    &\xmark&I (1.28M)  & 65.7 & 63.3 & 68.1 & 57.1 & 79.3 & 35.5 \\
DenseCL\cite{wang2021dense}  &\xmark&I (1.28M)  & 61.4 & 60.0 & 62.9 & 58.7 & 81.4 & 32.9 \\
PixPro\cite{xie2021propagate}   &\xmark&I (1.28M)  & 57.5 & 56.6 & 58.3 & 57.8 & 80.8 & 29.6 \\
\hdashline
Colorization$^{\star}$~\cite{vondrick2018tracking} &\xmark& Kinetics (800 hours) & 34.0 & 34.6 & 32.7 & 45.2 & 69.6 & - \\
CorrFlow$^{\star}$~\cite{lai2019self}  &\xmark&  OxUvA (14 hours)  & 50.3 & 48.4 & 52.2 & 58.5 & 78.8 & - \\
MAST$^{\star}$~\cite{lai2020mast}  &\xmark&  YT-VOS (5.58 hours) & 65.5 & 63.3 & 67.6 & - & - & - \\
TimeCycle~\cite{wang2019learning}   &\xmark&  VLOG (344 hours) & 48.7 & 46.4 & 50.0 & 57.3 & 78.1 & 28.9 \\
UVC~\cite{li2019joint}  &\xmark& Kinetics (800 hours) & 60.9 & 59.3 & 62.7 & 58.6  & 79.6 & 34.1 \\
CRW~\cite{jabri2020walk} &\xmark& Kinetics (800 hours) & 67.6 & 64.8 & 70.2 & 58.8 & 80.3 & 37.6 \\
\hdashline

FC(Ours) &\xmark&  YT-VOS (5.58 hours) & 67.7 & 64.7 & 70.5 & 59.3 & 80.8 & 34.0 \\
\textbf{SFC(Ours)} &\xmark&  YT-VOS, I(5.58 hours + 1.28M) & \textbf{71.2} & \textbf{68.3} & \textbf{74.0} & \textbf{61.9}  & \textbf{83.0} & 38.4     \\
\hline
OSVOS~\cite{caelles2017one} &\cmark   &I/D (1.28M + 10k) & 60.3 & 56.6 & 63.9 & - & - & - \\
OnAVOS~\cite{voigtlaender2017online} &\cmark  &I/C/P/D (1.28M + 517k)   & 65.4 & 61.6 & 69.1 & - & - & - \\
FEELVOS~\cite{voigtlaender2019feelvos}  &\cmark  &I/C/D/YT-VOS (1.28M + 663k)   & 71.5 & 69.1 & 74.0 & - & - & - \\
PAAP~\cite{iqbal2017pose} &\cmark & I/J (1.28M + 32K) & - & - & - & 51.6 & 73.8 & - \\
Thin-Slicing~\cite{song2017thin} &\cmark & I/J (1.28M + 32K) & - & - & - & 68.7 & 92.1 & - \\
ATEN~\cite{zhou2018adaptive} &\cmark & VIP (20k) & - & - & - & - & - & 37.9  \\
  \end{tabular}
}
\vspace{-2em}
\end{table*}

\subsubsection{Baselines} We compare with the following baselines:

\noindent \textit{Instance-Level Pre-Trained Representations:} We consider supervised and self-supervised pre-trained models (MoCo, BYOL, SimSiam, etc.) on ImageNet. We also compare with two recent video-based self-supervised representation learning baselines: VINCE~\cite{gordon2020watching} and VFS~\cite{xu2021rethinking}. We evaluate VFS pre-trained model using our label propagation implementation (official CRW~\cite{jabri2020walk} evaluation code).

\noindent \textit{Pixel-Level Pre-Trained Representations:} We evaluate representations trained with pixel-level self-supervised proxy tasks: PixPro\cite{xie2021propagate}, DetCo\cite{xie2021detco}, DenseCL\cite{wang2021dense}.

\noindent \textit{Task-Specific Temporal Correspondence Representations:} There are many self-supervised methods designed specifically for visual correspondence learning and evaluated on label propagation. We include these for a more comprehensive analysis: Colorization~\cite{vondrick2018tracking}, CorrFlow~\cite{lai2019self}, MAST~\cite{lai2020mast}, TimeCycle~\cite{wang2019learning}, UVC~\cite{li2019joint}, CRW~\cite{jabri2020walk}.

\subsection{Comparison with State-Of-The-Art}

We compare our method against previous self-supervised methods in Table~\ref{tab:davis}.
In summary, our results strongly validate the design choices in our model. In particular, the full semantic-aware fine-grained correspondence network (SFC), achieves state-of-the-art performance on all tasks investigated. SFC significantly outperforms other methods that learn only semantic correspondence (MoCo, $65.9 \rightarrow 71.2$ on DAVIS-2017) or only fine-grained correspondence (FC, $67.7 \rightarrow 71.2$ on DAVIS-2017). SFC even outperforms several supervised baselines specially designed for video object segmentation and human part tracking.

Note also that our fine-grained correspondence network (FC) can achieve comparable performance on DAVIS and JHMDB with methods like CRW, despite training with far less data and discarding temporal information. The performance of FC on VIP is lower, but it may be further improved by exploiting more inductive bias, e.g., temporal context or viewpoint changes in videos.

We show the results on DAVIS-2017 of FC using different pre-training datasets in Appendix~\ref{sec:diff_dataset}. FC achieves 67.9 $\mathcal{J\&F}_\textrm{m}$ when pre-trained on ImageNet. This suggests that a larger dataset offers marginal benefits for fine-grained correspondence learning, which is largely different from learning semantic correspondence. 
When replacing YouTube-VOS pre-trained FC with ImageNet pre-trained one, SFC still achieves 71.3 $\mathcal{J\&F}_\textrm{m}$. This indicates that the performance gain of SFC doesn't come from the extra YouTube-VOS dataset. We use YouTube-VOS for faster training and fair comparisons of other correspondence learning methods.

We also report results of SFC on \emph{semantic segmentation} and ImageNet-1K \emph{linear probing} in Appendix~\ref{sec:semantic_and_linear}. Our SFC achieves improved results on all considered tasks, showing strong generalization ability and the flexibility of our core contribution.

\begin{figure*}[t]
\begin{center}
\includegraphics[width=\linewidth]{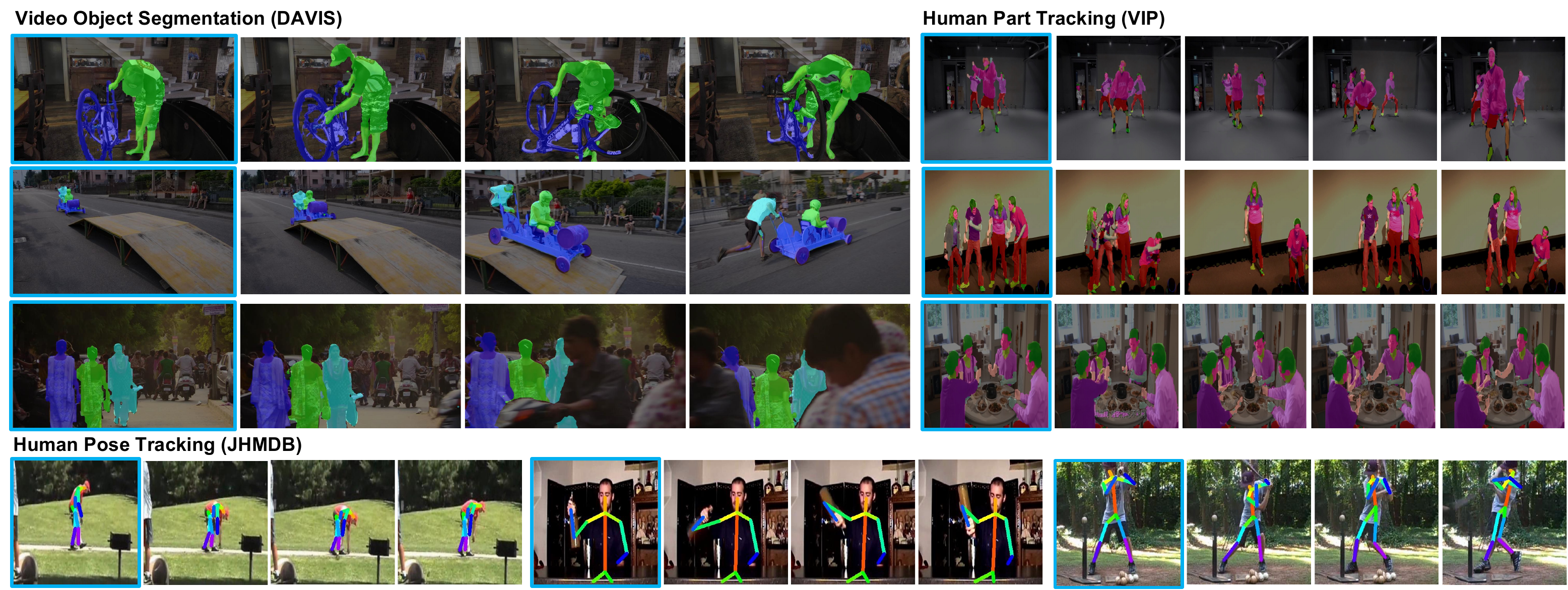}
\caption{\textbf{Qualitative results for label propagation.} Given ground-truth labels in the first frame (outlined in blue), our method can propagate them to the rest of frames. For more results, please refer to the Appendix~\ref{sec:propagation_vis}. }
\label{fig:mainvis}
\end{center}
\vspace{-3em}
\end{figure*}

\subsection{Visualization}

\begin{wrapfigure}{r}{0.45\textwidth}
\vspace{-4em}
\begin{center}
    \includegraphics[width=0.45\textwidth]{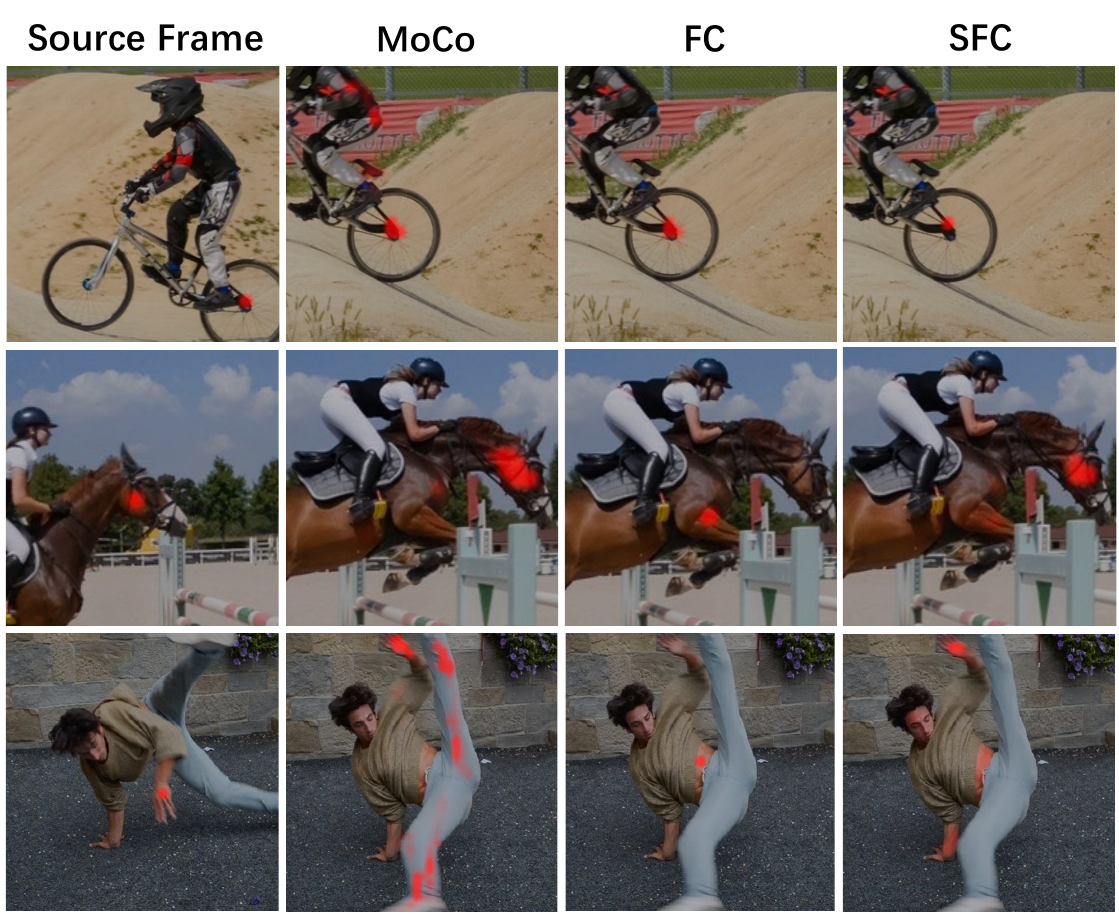}
\caption{\textbf{Correspondence map of SFC, compared with MoCo and FC.} Given the source frame with one pixel highlighted in red, we calculate the feature similarity between the target frame and this pixel. Red regions indicate high similarity.}
\label{fig:correspndence_vis}
\vspace{-2em}
\end{center}
\end{wrapfigure}

Figure~\ref{fig:mainvis} shows samples of video label propagation results. We further visualize the learned correspondences of our model in Figure~\ref{fig:correspndence_vis}, compared with its components, MoCo and FC. We notice that the correspondence map of MoCo tends to scatter across the entire visual object, indicating that it focuses more on object-level semantics instead of low-level fine-grained features. On the contrary, the correspondence map of FC is highly concentrated, but sometimes loses track of the source pixel, indicating a failure to capture high-level semantics. By balancing semantics and fine-grained correspondences, our proposed method SFC is able to overcome their respective drawbacks and give the most accurate correspondence.

\subsection{Ablative Analysis}
In this section, we investigate our results on video object segmentation using DAVIS-2017 in more detail, and outline several ablations on important design choices throughout our model architectures and pipelines.

\subsubsection{Fusion Strategy}
\label{sec:fusion_strategy}

We perform experiments by combing the FC training objective with a global image-level loss, resulting in an end-to-end multi-task framework. But we find the two losses fail to boost performance synergistically.
For example, when we add a BYOL loss to FC for joint optimization (see Appendix~\ref{sec:joint_training} for details), the performance on DAVIS-2017 drops a little (67.7 $\rightarrow$ 67.2). The reason is that the two losses need different receptive fields and augmentations. The optimal configuration of FC model will induce a sub-optimal solution under the image-level loss, and vice versa. Thus, it is sensible to train two independent models and use concatenation to fuse the two different kinds of representations.

\begin{wraptable}{r}{0.5\columnwidth}
\vspace{-3em}
\tablestyle{5pt}{1.1}
\caption{\textbf{Fusion of two networks with the same kind of correspondence.} FC denotes our fine-grained correspondence network, it achieves 67.7 $\mathcal{J}\&\mathcal{F}_\textrm{m}$ on DAVIS-2017. Other single model results can be found in Table~\ref{tab:davis}.}
\label{tab:combine_strategy}
\vspace{0.5em}
\adjustbox{width=\linewidth}{
\begin{tabular}{c|c|ccc}
    Type & Combination & $\mathcal{J}\&\mathcal{F}_\textrm{m}$ & $\mathcal{J}_\textrm{m}$  & $\mathcal{F}_\textrm{m}$ \\
    \shline
    \multirow{6}{*}{\tabincell{c}{Semantic\\Correspondence}} & InstDis + MoCo & 67.4 & 65.0 & 69.7 \\
                              & SimCLR + MoCo & 67.2 & 64.7 & 69.6 \\
                              & BYOL + MoCo & 67.5 & 65.0 & 70.0 \\
                              & Simsiam + MoCo & 67.4 & 65.1 & 69.6 \\
                              & VINCE + MoCo   & 66.7 & 64.2 & 69.2 \\
                              & VFS + MoCo     & 68.1 & 66.1 & 70.2 \\

    \hline
    \multirow{3}{*}{\tabincell{c}{Fine-grained\\Correspondence}} &TimeCycle + FC & 67.8 & 65.2 & 70.5 \\
                              & UVC + FC       & 61.5 & 59.8 & 63.3 \\
                              & CRW + FC  & 68.8 & 65.7 & 71.9 \\
\end{tabular}}
\vspace{-2em}
\end{wraptable}

One may expect the concatenation operation is some form of model ensemble. Does combining an arbitrary two networks lead to any reasonable improvement in performance? To answer this, we conduct experiments on two sets of models: in the first set, all models have two semantic correspondence networks; while in the second set, all models have two fine-grained correspondence networks. Results are shown in Table~\ref{tab:combine_strategy}. We observe that if two networks have the same type of correspondence, their combination leads to unremarkable increases in performance.

In Appendix~\ref{sec:add_exp}, we show that we can flexibly replace semantic correspondence backbone (MoCo $\rightarrow$ InstDis, SimCLR, BYOL, etc.) and still maintain strong performance on DAVIS. This strongly confirms our hypothesis that image-level self-supervised representations in general contain information about semantic correspondence. It also supports our framing of semantic correspondence and fine-grained correspondence as orthogonal sources of information.

\smallskip
Next, we mainly conduct a series of ablation studies on our fine-grained network (FC).

\subsubsection{Crop Size and Positive Radius}

\begin{wrapfigure}{r}{0.45\textwidth}
\vspace{-3em}
\begin{center}
\includegraphics[width=0.45\textwidth]{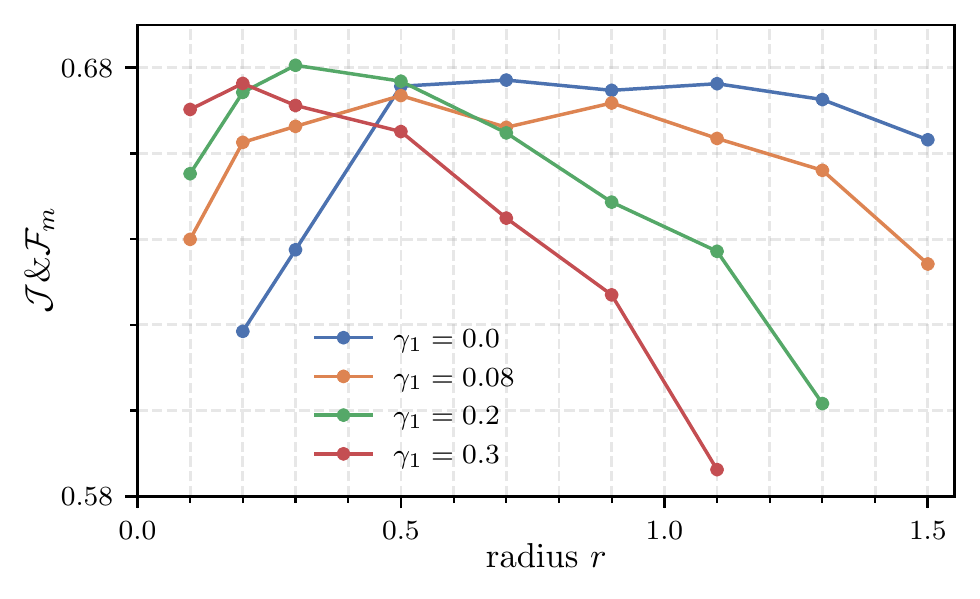}
\vspace{-2em}
\caption{\textbf{Effect of random crop size and positive radius.}}
\label{fig:crop_posratio}
\end{center}
\vspace{-3em}
\end{wrapfigure}

When we apply cropping for an image, a random patch is selected, with an area uniformly sampled between $\gamma_{1}$ (lower bound) and $\gamma_{2}$ (upper bound) of that of the original image.
In Figure~\ref{fig:crop_posratio}, we plot FC model performance on different ratios of crop size area, by varying $\gamma_{1}$: $\{0, 0.08, 0.2, 0.3\}$ and fixing $\gamma_{2}$ to 1. Simultaneously, for every lower bound $\gamma_{1}$, we investigate how different positive radii $r$ can also affect performance on correspondence learning.

We find that as the lower bound $\gamma_{1}$ increases, mode performance worsens. $\gamma_{1}=0$ yields relatively strong performance under a wide range of positive radius $r$. We conjecture that using a small lower bound $\gamma_{1}$ results in larger scale and translation variations between two views of one image, which induces strong spatial augmentation and thus allows our correspondence learning to rely on scale-invariant representations.
We also observe that an appropriate positive radius $r$ is crucial for fine-grained correspondence learning. On the DAVIS dataset, we show that a large (smooth) or small (sharp) $r$ is demonstrably harmful to performance. Finally, for different $\gamma_{1}$, the optimal value of $r$ is different.

\subsubsection{Data Augmentation}

\begin{wraptable}{r}{0.45\columnwidth}
\vspace{-3em}
\centering
\small
\tablestyle{8pt}{1.1}
\caption{\textbf{Effect of data augmentation.} The augmentation strategy follows MoCo v2~\cite{chen2020improved}.}
\label{tab:augmentation}
\vspace{0.5em}
\adjustbox{width=\linewidth}{
\begin{tabular}{l|ccc}
    Augmentation & $\mathcal{J}\&\mathcal{F}_\textrm{m}$ & $\mathcal{J}_\textrm{m}$  & $\mathcal{F}_\textrm{m}$ \\
    \shline
     Random crop                             & 67.6 & 64.7 & 70.5 \\
     Random crop \& Random flip              & 67.7 & 64.8 & 70.6 \\
     Random crop \& Color jittering          & 65.5 & 62.6 & 68.4 \\
     Random crop \& Gaussian blur            & 65.9 & 63.2 & 68.6 \\
     Random crop \& Color dropping           & 61.9 & 58.9 & 64.9 \\
\end{tabular}}
\vspace{-1em}
\end{wraptable}

VFS~\cite{xu2021rethinking} has pointed out that color augmentation jeopardizes fine-grained correspondence learning. To systematically study the effects of individual data augmentations, we investigate the performance of our FC model on DAVIS when applying random cropping and another common augmentation (random flip, color jittering, etc.). We report the results in Table~\ref{tab:augmentation}. Among all color data augmentations, the one that has the greatest negative impact on fine-grained correspondence learning is actually color dropping (grayscale conversion). This is in contrast to image-level self-supervised learning, where strong color augmentation~\cite{chen2020simple} is crucial for learning good representations. We adopt random crop as the only augmentation in our best-performing models.

\section{Conclusion and Discussions}

We have developed a novel framework to learn both semantic and fine-grained correspondence from still images alone. We demonstrate that these two forms of correspondence offer complementary information, thereby facilitating a simple yet intuitive fusion scheme which leads to state-of-the-art results on a number of downstream correspondence tasks. 
In this work, we mainly explore the correspondence properties of ConvNet. Whether ViT~\cite{dosovitskiy2020image} also benefits from dense fine-grained self-supervision and combination of two kinds of correspondence is an interesting open question left to future exploration.

\subsection*{Acknowledgements.}
This work is supported by the Ministry of Science and Technology of the People's Republic of China, the 2030 Innovation Megaprojects ``Program on New Generation Artificial Intelligence'' (Grant No. 2021AAA0150000).
This work is also supported by a grant from the Guoqiang Institute, Tsinghua University.

\clearpage
%
%
\bibliographystyle{splncs04}
\bibliography{egbib}

\clearpage

\renewcommand\thefigure{\thesection.\arabic{figure}}
\renewcommand\thetable{\thesection.\arabic{table}}
\setcounter{figure}{0} 
\setcounter{table}{0}

\appendix

\section{Implementation Details}

\subsection{FC pre-training}
\label{sec:fc_details}
The implementation details of our fine-grained correspondence network are as follows.

\noindent \textbf{Data Augmentation}\quad 
We use only spatial augmentation, where two random crops with scale $[0.0, 1.0]$ from the image are generated and resized into $256\times256$. 

\noindent \textbf{Architectures}\quad Following ~\cite{jabri2020walk,li2019joint,wang2019learning}, we adopt ResNet-18 as the backbone $\mathbf{f}$ and reduce the stride of last two residual blocks ($\texttt{res3}$ and $\texttt{res4}$) to $1$. The modified backbone produces a feature map with size $32 \times 32$ (ablation in Appendix~\ref{sec:feature_resolution}). The dense projection and prediction head use the same architecture: a $1 \times 1$ convolution layer with $2048$ output channels followed by batch normalization and a ReLU activation, and a final $1 \times 1$ convolution layer with output dimension $256$. The positive radius $r$ used to control the size of spatial neighborhood is set to $0.5$.

\noindent \textbf{Optimization}\quad We train the model with the Adam optimizer for 60k iterations. The learning rate is set to $0.001$. The weight decay is set to 0. The batch size is 96. For the target network, the exponential moving average parameter $\tau$ starts from 0.99 and gradually increases to 1 under a cosine schedule, following~\cite{grill2020bootstrap}. The whole model can be trained on a single $24$GB NVIDIA 3090 GPU.

\subsection{Label Propagation}
\label{sec:label_prop_detail}
We follow the same label propagation algorithm in \cite{jabri2020walk}. Specifically, given the ground-truth labels in the first frame, a recurrent inference strategy is applied to propagate the labels to the rest of the frames: we calculate the similarity between the current frame with the first frame (to provide ground truth labels) as well as the preceding $m$ frames (to provide predicted labels). We reduce the stride of the penultimate residual block ($\texttt{res4}$) of the backbone network to be 1 and use its output (stride 8) to compute a dense similarity matrix. To avoid ambiguous matches, we define a localized spatial neighborhood by computing the similarity between pixels that are at most ${r}'$ pixels away from each other. Finally, the labels of the top-$k$ most similarly local feature vectors are selected and are propagated to the current frame.

\label{sec:test_parameter}
For a single network which only learns semantic correspondence or fine-grained correspondence, the detailed test hyper-parameters for the three datasets are listed in Table~\ref{tab:hyper_single}.

\begin{table}[h]
\centering
\small
\tablestyle{3pt}{1.1}
\caption{Test hyper-parameters for a single network.
\label{tab:hyper_single}}
\begin{tabular}{l|ccc}
                        & DAVIS & JHMDB & VIP       \\
\shline
top-$k$                   & 10    & 10  & 10        \\
preceding frame $m$       & 20    & 8   & 8         \\
propagation radius $r'$   & 12    & 3   & 15        
\end{tabular}

\end{table}

Recall that in fusing the two different kinds of correspondence, we introduce a new hyper-parameter $\lambda$. We report the test hyper-parameters for combined correspondence in Table~\ref{tab:hyper_combine}. In general, we find that more neighbors (larger top-k and propagation radius $r'$) are required for consistent performance.

\begin{table}[h]
\centering
\small
\tablestyle{3pt}{1.1}
\caption{Test hyper-parameters when fusing two kinds of correspondence.
\label{tab:hyper_combine}}
\begin{tabular}{l|ccc}
                        & DAVIS & JHMDB & VIP    \\
\shline
weight $\lambda$        & 1.75  & 1.0 & 1.0       \\
top-$k$                 & 15    & 20  & 10        \\
preceding frame $m$     & 20    & 8   & 8         \\
propagation radius $r'$ & 15    & 5   & 15    
\end{tabular}
\end{table}

\subsection{Semantic Segmentation Protocol}
The backbone is kept fixed and we train a $1 \times 1$ convolutional layer on top to predict a semantic segmentation map. We apply dilated convolutions in the last residual block to obtain dense predictions. We use PASCAL\cite{everingham2010pascal} $\texttt{train\_aug}$ and $\texttt{val}$ splits during training and evaluation, respectively. We adopt mIoU as the metric. The $1 \times 1$ convolutional layer training uses base $lr=0.1$ for 60 epochs, weight decay $=0.0001$, momentum $=0.9$, and batch size $=16$ with an SGD optimizer.

\subsection{Linear Classification Protocol}
Given the pre-trained network, we train a supervised linear classifier on top of the frozen features, which are obtained from ResNet's global average pooling layer. We train this classifier on the ImageNet train set and report top-1 classification accuracy on the ImageNet validation set. Following prior work\cite{he2020momentum}, the linear classifier training uses base $lr=30.0$ for 100 epochs, weight decay $=0$, momentum $=0.9$, and batch size$=256$ with a SGD optimizer.

\subsection{Combined with Image-level Pretext Task}
\label{sec:joint_training}
We add BYOL loss to FC for joint optimization. Specifically, the two loss functions share the same backbone encoder (outputs a feature map with size 32 $\times$ 32) and data loader (performs only spatial augmentation). But the projection head and prediction head are not shared. The projection head of BYOL is a two-layer MLP whose hidden and output dimensions are 2048 and 256. Note that BYOL average-pool backbone features to aggregate information from all spatial locations. Other implementation details follow FC. Two loss functions are balanced by a multiplicative factor $\alpha$ (set to 1 by default).

\section{Additional Experimental Results}
\label{sec:add_exp}

\subsection{FC is Robust to Different Dataset}
\label{sec:diff_dataset}
When pretrained on non object-centric dataset (e.g. COCO~\cite{lin2014microsoft}), the performance of typical image-level self-supervised methods drop significantly~\cite{purushwalkam2020demystifying,selvaraju2021casting}. At the same time, it is largely recognized that a larger dataset usually results in stronger semantic representation for these methods. But this may not be true for a task that requires analyzing low-level cues. 
The following Table~\ref{tab:diff_dataset} compares different training datasets of FC. We can see that FC is robust to the size and nature of the dataset. FC can effectively learn from a relatively small dataset. It actually gains more benefits from datasets that contain more complex scenes with several objects. The results on COCO even surpass Youtube-VOS used in the main body of the paper.

\begin{table}[t]
\centering
\small
\tablestyle{5pt}{1.1}
\caption{\textbf{Results on DAVIS-2017 of FC using different training datasets.} The number of images per dataset is in parentheses.}
\label{tab:diff_dataset}
\begin{tabular}{c | c c c c}
Dataset & PASCAL(17K) & COCO(118K) &YT-VOS(95K) & ImageNet(1.28M) \\
\shline
$\mathcal{J}\&\mathcal{F}_\textrm{m}$ & 67.9 & 68.2 & 67.7 & 67.9  \\
\end{tabular}
\end{table}

\subsection{Feature Resolution}
\label{sec:feature_resolution}
We report the results of our fine-grained correspondence network (FC) using different feature resolutions in Table~\ref{tab:resolution}. The performance on DAVIS improves as the resolution increases. This is intuitive, because higher resolution indicates the local feature vectors correspond to a smaller region on the original image (small receptive fields), which benefits fine-grained low-level correspondence learning. But high-level semantics require larger receptive fields to encode more holistic information.

\begin{table}[t]
\centering
\small
\tablestyle{4pt}{1.1}
\caption{\textbf{Effect of feature map resolution.} The results increase as resolution gets higher. We use 32 $\times$ 32 by default.}
\label{tab:resolution}
\begin{tabular}{c|ccc}
    Feature Resolution & $\mathcal{J}\&\mathcal{F}_\textrm{m}$ & $\mathcal{J}_\textrm{m}$  & $\mathcal{F}_\textrm{m}$ \\
    \shline
    8  $\times$ 8                 & 63.3 & 61.8 & 64.8 \\
    16 $\times$ 16                & 65.2 & 63.4 & 67.0 \\
    32 $\times$ 32                & 67.6 & 64.7 & 70.5 \\
\end{tabular}
\end{table}

\subsection{Semantic Segmentation and Linear Classification}
\label{sec:semantic_and_linear}

The quantitative comparison on different downstream tasks is shown in Table~\ref{tab:segmentation_and_classification}. For a fair comparison, we use ResNet-18 as MoCo backbone. CRW surpasses MoCo on DAVIS, but is dramatically outperformed by MoCo on semantic segmentation and image classification. Note that our FC model exhibits similar properties as CRW: the learned representation is suitable for fine-grained correspondence task, but lacks high-level semantic information. When we add crucial missing semantic information, our SFC achieves significant improvements on label propagation, semantic segmentation and image classification.

\begin{table}[h]
\centering
\small
\tablestyle{5pt}{1.1}
\caption{Comparison on label propagation, semantic segmentation and linear classification.}
\label{tab:segmentation_and_classification}
   \begin{tabular}{c|ccc|c|c}
   & \multicolumn{3}{c|}{DAVIS}  & PASCAL & ImageNet\\
   Method & $\mathcal{J\&F}_\textrm{m}$ & $\mathcal{J}_\textrm{m}$ & $\mathcal{F}_\textrm{m}$ & mIoU  & Acc@1 \\
  \shline
  MoCo & 62.1 & 60.3 & 63.8 & 25.5 & 48.7 \\
  CRW  & 67.6 & 64.8 & 70.2 & 13.0 & 12.6 \\
  FC   & 67.7 & 64.7 & 70.5 & 15.9 & 16.3 \\
  SFC  & 69.5 & 66.7 & 72.4 & 30.1 & 51.2 \\
  \end{tabular}
\end{table}

\subsection{Semantic Correspondence Backbone}
\label{sec:semantic_backbone}

We use MoCo as the default semantic correspondence backbone in the main experiment, but our framework is extensible to any arbitrary backbone that is capable of producing spatial feature maps. In Table~\ref{tab:semantic_backbone}, we show that we can flexibly swap out the semantic correspondence backbone for any off-the-shelf self-supervised network and maintain strong performance on DAVIS. Some methods such as SimCLR and BYOL even surpass MoCo. This strongly supports our hypothesis that image-level representations in general contain information about semantic correspondences.

\begin{table}[h]
\centering
\small
\tablestyle{5pt}{1.1}
\caption{Results after replacing MoCo with alternate image-level self-supervised representation learning methods.}
\label{tab:semantic_backbone}
\begin{tabular}{l|ccc}
    Combination & $\mathcal{J}\&\mathcal{F}_\textrm{m}$ & $\mathcal{J}_\textrm{m}$  & $\mathcal{F}_\textrm{m}$ \\
    \shline
    InstDis + FC & 70.1 & 67.1 & 73.1 \\
    MoCo + FC    & 71.2 & 68.3 & 74.0 \\
    SimCLR + FC  & 71.4 & 68.6 & 74.2 \\
    BYOL + FC    & 71.3 & 68.4 & 74.1 \\
    SimSiam + FC & 70.2 & 67.3 & 73.1 \\
    VINCE + FC   & 70.4 & 67.7 & 73.2 \\
    VFS + FC     & 70.7 & 67.8 & 73.7 \\
\end{tabular}
\end{table}

\subsection{Fine-Grained Correspondence Backbone}
\label{sec:fine_grained_backbone}

In Table~\ref{tab:semantic_crw}, we replace our own FC network in SFC with another fine-grained correspondence network CRW. We find the performance generally underperforms SFC. FC is better than CRW on all evaluation metrics, as shown in Table~\ref{tab:segmentation_and_classification}. FC is also much simpler and computationally efficient. It takes less than a day using a single GPU, but CRW reports seven days of training. 

The results in Table~\ref{tab:semantic_crw} surpass image-level self-supervised methods or CRW alone, demonstrating the benefits of considering two orthogonal correspondences and the flexibility of our framework. It enables us to explore more effective and efficient self-supervised learning methods for semantic or fine-grained representations separately.

\begin{table}[h]
\centering
\small
\tablestyle{5pt}{1.1}
\caption{Replace FC with another fine-grained correspondence model CRW.}
\label{tab:semantic_crw}
\begin{tabular}{l|ccc}
    Combination & $\mathcal{J}\&\mathcal{F}_\textrm{m}$ & $\mathcal{J}_\textrm{m}$  & $\mathcal{F}_\textrm{m}$ \\
    \shline
    InstDis + CRW & 69.6 & 66.6 & 72.6 \\
    MoCo + CRW    & 70.6 & 67.8 & 73.4 \\
    SimCLR + CRW  & 70.7 & 68.0 & 73.4 \\
    BYOL + CRW    & 70.9 & 68.1 & 73.6 \\
    SimSiam + CRW & 69.7 & 66.8 & 72.6 \\
    VINCE + CRW   & 70.3 & 67.6 & 73.1 \\
    VFS + CRW     & 70.6 & 67.7 & 73.5 \\
\end{tabular}
\end{table}

\section{Visualization}
\label{sec:propagation_vis}
\begin{figure*}[t]
\begin{center}
\includegraphics[width=\linewidth]{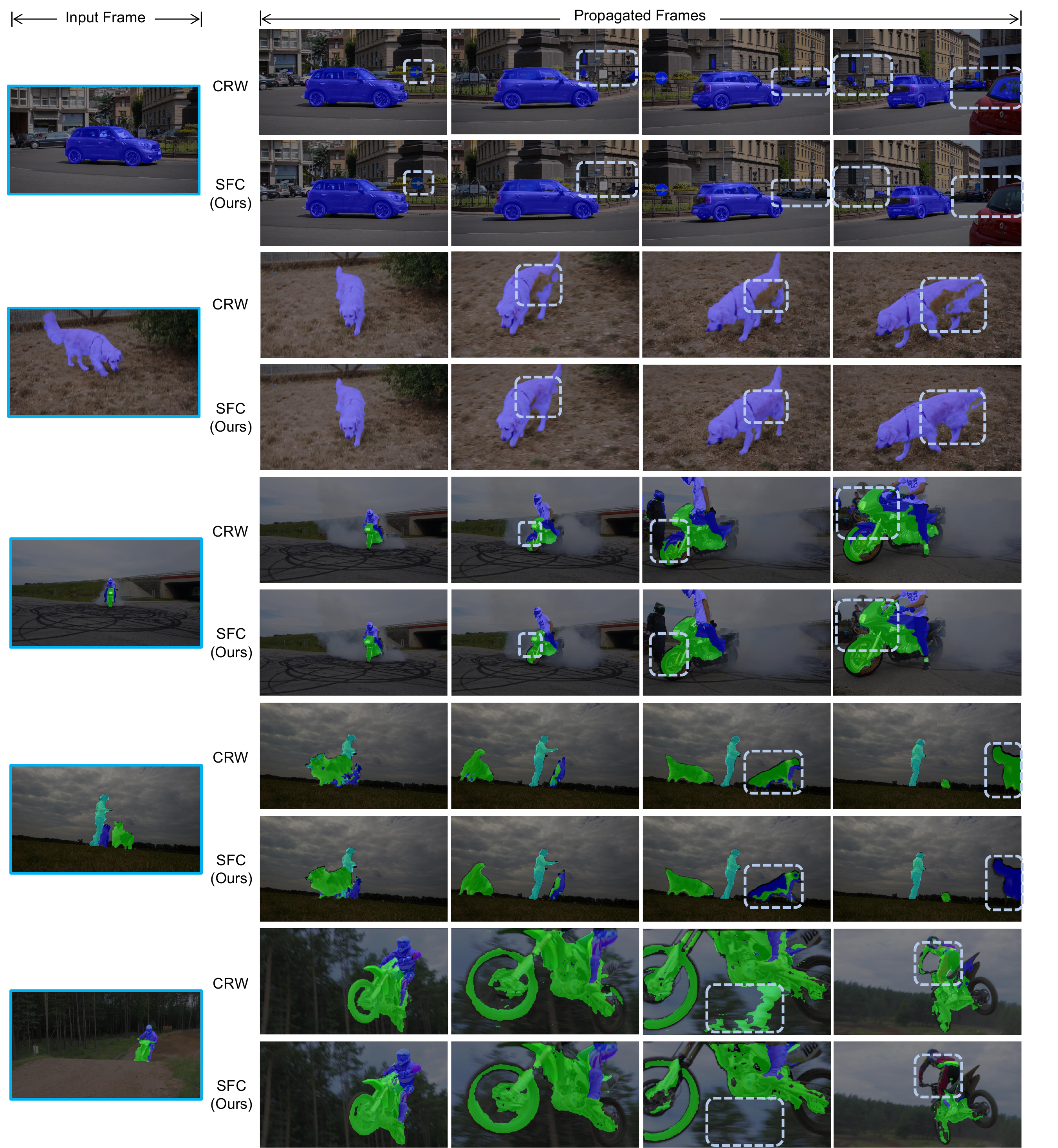}
\caption{\textbf{Comparing our SFC with CRW on DAVIS-2017.} Within each example, the upper row is the output of CRW, and the lower row is the output of SFC. Blue dashed boxes indicate the main areas of difference.}
\label{fig:detailed_vis}
\end{center}
\end{figure*}

\begin{figure*}[t]
\begin{center}
\includegraphics[width=\linewidth]{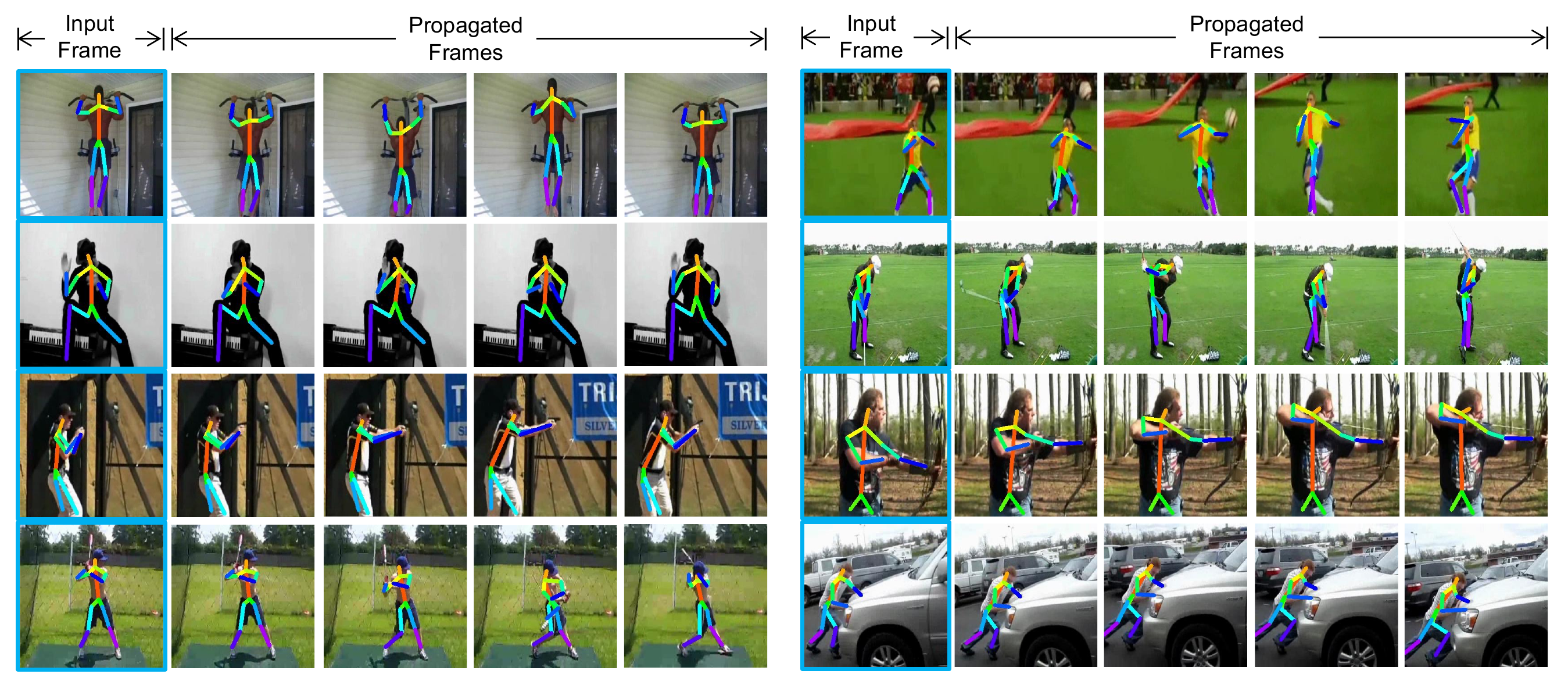}
\caption{\textbf{Visualization on JHMDB.} Pose keypoints and their initial positions are defined on the input frame (outlined in blue), and propagated to the rest of frames.}
\label{fig:detailed_vis_2}
\end{center}
\vskip -0.3 in
\end{figure*}

\begin{figure*}[t]
\begin{center}
\includegraphics[width=\linewidth]{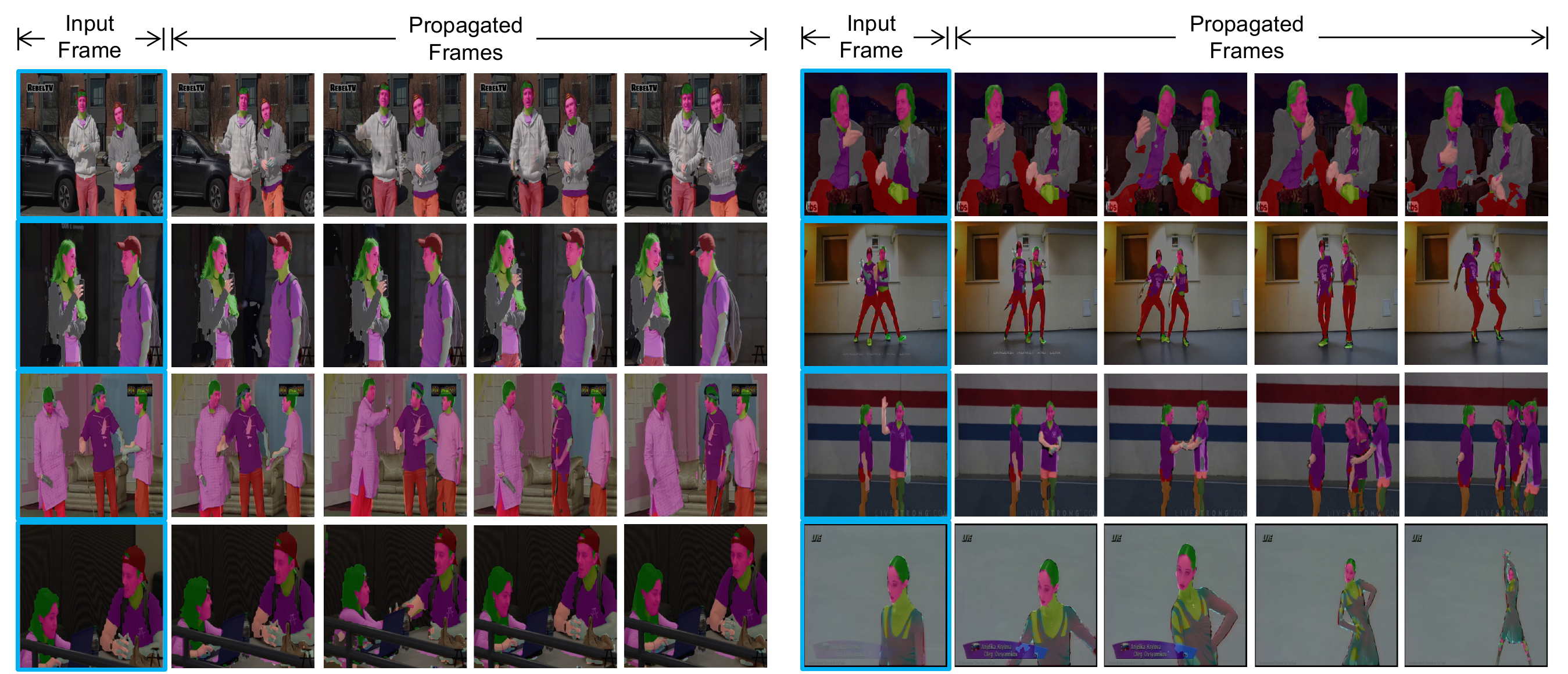}
\caption{\textbf{Visualization on VIP.} The segmentation map of different body parts are defined on the input frame (outlined in blue), and propagated to the rest of frames.}
\label{fig:detailed_vis_3}
\end{center}
\vskip -0.3 in
\end{figure*}

We provide a more detailed visualization of our SFC model on several downstream label propagation tasks. 
In Figure~\ref{fig:detailed_vis}, we show a comparison between SFC and CRW on the visual object segmentation benchmark DAVIS-2017. Our SFC model can generally output more accurate segmentation boundaries and reduce the amount of mistakes and failures made by CRW.  
In Figure~\ref{fig:detailed_vis_2} and Figure~\ref{fig:detailed_vis_3}, we provide visualizations on the human pose tracking benchmark JHMDB and the human part tracking benchmark VIP. Note that in all our experiments, no prior knowledge on human structure or object class is used. The label propagation process is solely based on feature matching.

\end{document}